\documentclass[conference]{IEEEtran}
\IEEEoverridecommandlockouts
\usepackage{cite}
\usepackage{amsmath,amssymb,amsfonts}
\usepackage{algorithmic}
\usepackage{graphicx}
\usepackage{graphicx,subcaption}
\usepackage{authblk}
\usepackage{textcomp}
\usepackage{xcolor}
\def\BibTeX{{\rm B\kern-.05em{\sc i\kern-.025em b}\kern-.08em
    T\kern-.1667em\lower.7ex\hbox{E}\kern-.125emX}}
\begin{document}

\title{xMTrans: Temporal Attentive Cross-Modality Fusion Transformer for Long-Term Traffic Prediction\\

}

\author[1]{Huy Quang Ung}
\author[1]{Hao Niu}
\author[2]{Minh-Son Dao}
\author[3]{Shinya Wada}
\author[1]{Atsunori Minamikawa}
\affil[1]{KDDI Research, Inc., Saitama, Japan}
\affil[2]{National Institute of Information And Communications Technology, Tokyo, Japan}
\affil[3]{KDDI Corp., Iidabashi, Japan}
\affil[ ]{xhu-ung@kddi.com, ha-niu@kddi.com, dao@nict.go.jp, sh-wada@kddi.com, at-minamikawa@kddi.com}
% 

% \author{\IEEEauthorblockN{1\textsuperscript{st} Huy Quang Ung}
% \IEEEauthorblockA{\textit{Human-centered AI Laboratory} \\
% \textit{KDDI Research, Inc.\\
% Fujimino, Japan \\
% xhu-ung@kddi.com}}
% \and
% \IEEEauthorblockN{2\textsuperscript{nd} Hao Niu}
% \IEEEauthorblockA{\textit{Human-centered AI Laboratory} \\
% \textit{KDDI Research, Inc.}\\
% Fujimino, Japan \\
% ha-niu@kddi.com}
% \and
% \IEEEauthorblockN{3\textsuperscript{rd} Minh-Son Dao}
% \IEEEauthorblockA{\textit{Big Data Integration Research Center} \\
% \textit{National Institute of Information and Communications Technology (NICT)}\\
% Koganei, Japan \\
% dao@nict.go.jp}
% \and
% \IEEEauthorblockN{4\textsuperscript{th} Shinya Wada}
% \IEEEauthorblockA{\textit{Data and AI Center} \\
% \textit{KDDI Corporation}\\
% Iidabashi, Japan \\
% sh-wada@kddi.com}
% \and
% \IEEEauthorblockN{5\textsuperscript{th} Atsunori Minamikawa}
% \IEEEauthorblockA{\textit{Human-centered AI Laboratory} \\
% \textit{KDDI Research, Inc.}\\
% Fujimino, Japan \\
% at-minamikawa@kddi.com}

% \and

\maketitle

\begin{abstract}
Traffic predictions play a crucial role in intelligent transportation systems. 
The rapid development of IoT devices allows us to collect different kinds of data with high correlations to traffic predictions, fostering the development of efficient multi-modal traffic prediction models.
Until now, there are few studies focusing on utilizing advantages of multi-modal data for traffic predictions.
In this paper, we introduce a novel temporal attentive cross-modality transformer model for long-term traffic predictions, namely \emph{xMTrans}, with capability of exploring the temporal correlations between the data of two modalities: one target modality (for prediction, e.g., traffic congestion) and one support modality (e.g., people flow). 
We conducted extensive experiments to evaluate our proposed model on traffic congestion and taxi demand predictions using real-world datasets. 
The results showed the superiority of \emph{xMTrans} against recent state-of-the-art methods on long-term traffic predictions. 
In addition, we also conducted a comprehensive ablation study to further analyze the effectiveness of each module in \emph{xMTrans}.
\end{abstract}


\def \tp {TP}
\def \lttp {LTTP}
\def \sm {SM}
\def \tm {TM}

\section{Introduction}

Transportation and traveling are essential activities of human beings in their daily life. 
Following rapid expansion of our civilization, the demands of traffic usage are significantly increasing, especially in populous countries.
Consequently, typical problems in traffic management have been observed such as traffic congestion and public traffic overload in rush hours and special events. 
Long-term predicting the occurrences of these problems could assist traffic managers in adjusting traffic flow for prevention.
Besides, integrating these predictions into end-user navigation network systems is a possible solution to optimize the traffic flow by suggesting alternative routes to traffic participants.
In addition, exploring traffic congestion-prone regions and the underlying issues can provide insights for future urban development.

The traffic problems could be consequences of various factors, e.g., weather conditions, disasters, social events, and incidents.
Due to the popularity of IoT devices, these factors can be measured and utilized to assist the development of traffic applications.
When data related to these factors is available, utilizing it to develop multi-modal traffic prediction (\tp) models is a promising approach~\cite{chou2019long}.
Recently, our work~\cite{dao2023fostering} introduced a data platform, namely Data Centric Cloud Service (DCCS)~\footnote{https://testbed.nict.go.jp/dccs/}, consisting of multi-modal dataset (i.e., traffic congestion length, weather, rainfall amount, and people-flow data) for benchmarking urban risk transportation prediction. 
Those datasets will be publicly available to the research community in the near future.
%We use a part of this multi-modal datasets in this study's experiments.

Among those factors, people-flow data collected from mobile phone location under user's consent agreement could be a useful feature for \tp~models. 
Especially, capturing the temporal correlations between traffic problem and people-flow is necessary for these models. 
For instance, the volume of people in a certain area is high at a moment since a crowded event is being held, then there might be a traffic congestion or high traffic demand several hours later when the event is ended. 

In addition, there are also correlations among different transportation methods. 
For instance, when the demand of bus usage in a certain location is suddenly increasing, there could be a possibility of an increasing taxi demand due to the overload of buses. 
Notably, Ye et al.~\cite{ye2019co} showed that a co-prediction model of bike and taxi demands achieved better performance than individual models which are only trained by their own domain data. 
It, therefore, indicates the efficacy of multi-modal modelling in \tp.

Up to now, there are few research studies tackling on multi-modal modelling for long-term \tp~(\lttp).
Chou et al.~\cite{chou2019long} integrated the weather effect into a Long-Short Term Memory based model for long-term traffic time prediction.
Balsam et al.~\cite{alkouz2020snsjam} leveraged textual cross-lingual data from social networks to improve the performance of their linear regression \tp~model.
Similarly, Dao et al.~\cite{3DCNN} introduced an early fusion 3DCNN model incorporating social network content related to traffic issues such as crash, heavy rain, congestion, etc.
Furthermore, Dao et al.~\cite{dao2021insights} extended their work that added rainfall data to their proposed model for performance improvement.
However, these studies applied simple aggregation or linear transformation methods to combine historical traffic data and the auxiliary/support modality data.
It, therefore, is lack of deeply exploiting correlations among the data of different modalities. 
This is still an open challenge to be solved.

In general, \tp~can be treated as a multivariate time-series prediction (MTSP) task. Because of the superiority of capturing long-term dependencies, Transformer~\cite{vaswani2017attention} has become a popular structure for the long-term MTSP. 
The recent Transformer-based PatchTST~\cite{2023_nie_patchtst} model outperformed other previous methods in long-term MTSP including \tp, showing the effectiveness of Transformer structure in traffic modelling. 
However, these methods are not designed to leverage advantages of multi-modal data.
In this paper, we adopt the Transformer structure to effectively exploit the temporal correlations of multi-modal data. 
Specifically, we propose a novel cross-modality Transformer model (denoted as \emph{xMTrans}) with capability of exploring the correlations along the time dimension between the data of two modalities, i.e., a target modality (for prediction, denoted as \tm) and a support modality (denoted as \sm).
In our cross-modality fusion block, we utilize multiple layers of masked multi-head attention to fuse multi-modal data.
Along with the masked multi-head self-attention module, we introduce a masked multi-head temporal attention module to retrieve values of support modality at correlated timestamps by time-formed keys and queries. 
In addition, we use the multi-resolution recursive training strategy in~\cite{niu2022mu2rest} to enhance the performance of the proposed architecture.

Our contributions are listed as follows:
\begin{itemize}
    \item We introduce a novel cross-modality fusion transformer, i.e., \emph{xMTrans}, for \lttp. 
    % To the best of our knowledge, this is the first multi-modal LTTP model following the intermediate fusion approach.
    \item We show the effectiveness of \emph{xMTrans} on traffic congestion length and taxi demand prediction tasks using real-world datasets.
    For both tasks, using an additional mobility data as the \sm~is able to significantly improve the prediction performance.
    \item We compare our proposed method to several recent state-of-the-art LTTP models and show the superiority against them. In addition, we conduct a detailed ablation study to evaluate different modules in \emph{xMTrans}.
\end{itemize}

The remaining of this paper is organized as follows. 
Section 2 presents related work on long-term traffic prediction. 
Section 3 describes the problem formulation. 
Section 4 is to introduce our proposed model in detail. 
Evaluations of our model are presented in Section 5. 
Finally, Section 6 concludes this work and discusses future works. 

\section{Related Work}

Up to now, these are still few studies focusing on multi-modal TP. 
Han et al.~\cite{han2021dynamic} utilized traffic volume as an auxiliary feature to predict short-term traffic speed using a dynamic graph neural network. 
In addition, Ye et al.~\cite{ye2019co} proposed a heterogeneous Long-Short Term Memory (LSTM) based model for short-term co-prediction of bike and taxi demands using external factors such as weather, event. 
Those factors are embedded into a feature space and added directly to hidden states of the LSTM model.

Many new models are emerging recently for the long-term multivariate time-series prediction (MTSP) including the TP task. Most of these models are based on the Transformer~\cite{vaswani2017attention} or Linear/Multi-layer Perception (MLP) layer. The Transformer based models, such as
Informer~\cite{2021_zhou_informer}, 
Autoformer~\cite{2021_wu_autoformer}, Pyraformer~\cite{2022_liu_pyraformer} and FEDformer~\cite{2022_zhou_fedformer}, take the advantage of the Transformer in capturing long-term dependencies among the tokens. 
To reduce the complexity of applying Transformer for MTSP, Informer proposed ProbSparse self-attention mechanism and self-attention distilling operation, followed by a one-step forward procedure. 
Autoformer introduced time series decomposition in the Transformer structure and realizes a better prediction performance. 
Pyraformer built the pyramidal attention module to utilize the multi-resolution information of the data and FEDformer considered the attention in the frequency domain.
More recently, a study~\cite{zeng2023transformers} using a simple Linear layer appeared to doubt the effectiveness of the Transformer-based models for MTSP. 
In that study, three Linear-based models (Linear, NLinear and DLinear) with channel-independent (CI) training strategy are proposed to beat these Transformer-based models. 
By adopting the CI training strategy and an instance normalization method (RevIN)~\cite{kim2022reversible}, PatchTST illustrated the priority of the Transformer structure again. 
The work~\cite{li2023revisiting} further proposed Linear/MLP-based models (RLinear and RMLP) with RevIN which are able to achieve competitive performance to the PatchTST. 
% A recent work of iTransformer~\cite{liu2023itransformer} inverted the structure of Transformer' architecture in which it treats a variate as a token, achieving the best performance on common benchmarks.
However, all of these recent works focus on uni-modal data without considering the efficient use of multi-modal data.

\section{Problem Formulation}
We formulate the LTTP problem as long-term MTSP. 
Particularly, we focus on the traffic congestion length and taxi demand prediction in this paper.
Let's assume that we have two modalities: (1) a target modality (TM) aiming to predict, (2) a support modality (SM) used as a auxiliary feature for our TM prediction. 
Given historical records of $(H+1)$ time steps ($H\geq 0$) from $N$ locations, the task of the model is to predict the future $L$ time steps ($L\geq1$) for each location, i.e., predict $[X_{(t+1)},\dots,X_{(t+L)}]_{n}$ of \tm~given $[X_{(t-H)},\dots,X_t]_{n}$ of \tm~and $[X'_{(t-H)},\dots,X'_t]_{n}$ of \sm, $n\in\{1, 2, ..., N\}$.
In addition, temporal features consisting of month, date, hour, minute, day of week, and holiday are available to use in both training and inference phases.
$R$ (e.g., 15 minutes) denotes a time interval between two consecutive time steps.

\begin{figure*}[t]
    \centering
    \includegraphics[scale=0.550]{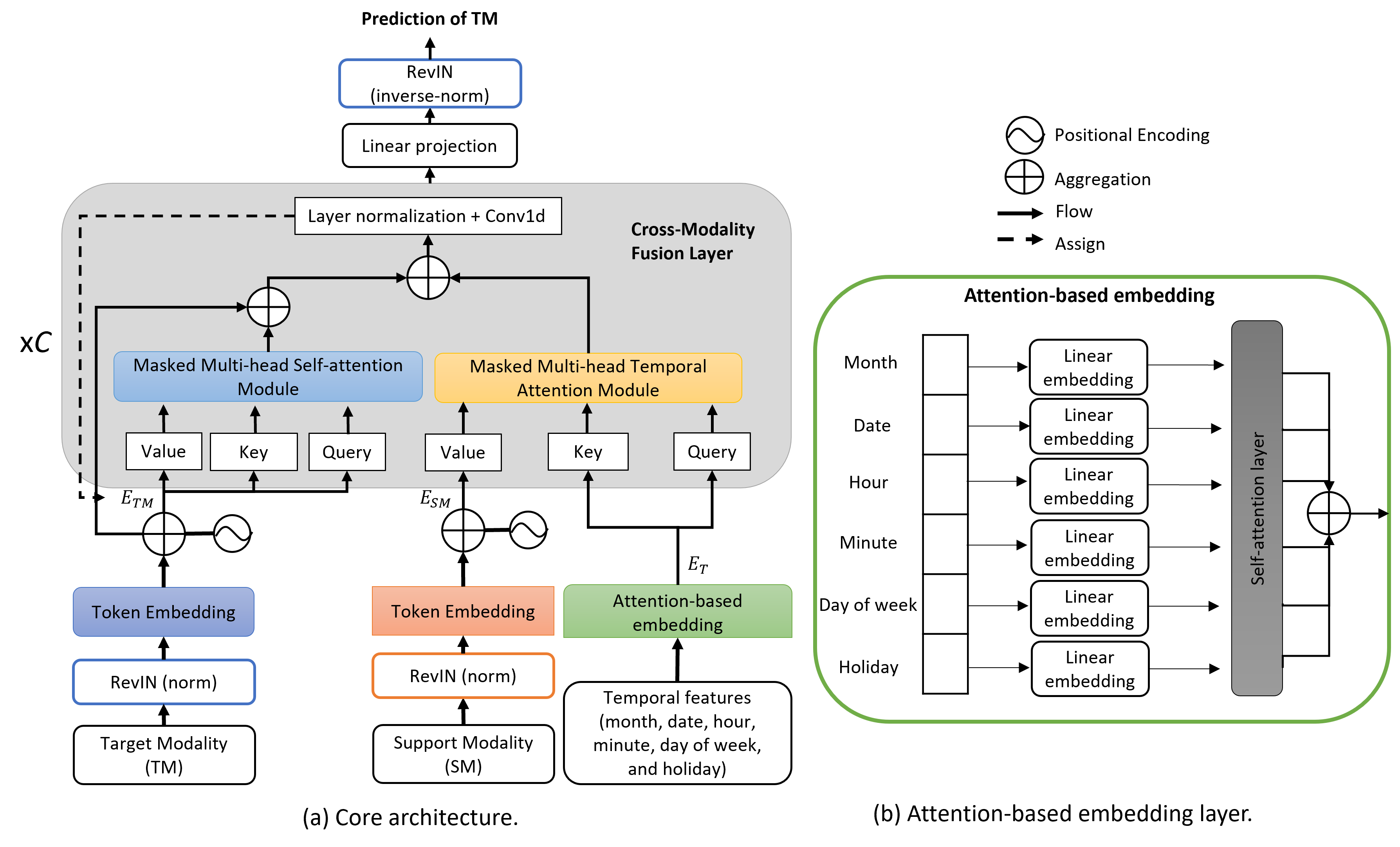}
    \caption{Overview of our proposed architecture.}
    \label{fig:overview}
\end{figure*}

\section{Proposed Method}
We introduce \emph{xMTrans}, a cross-modality temporal attentive fusion transformer for multivariate long-term traffic prediction.
The overview of our proposed architecture is presented in Fig.~\ref{fig:overview}.
% Similar to recent LTTP methods (e.g., PatchTST~\cite{2023_nie_patchtst}, RLinear, RMLP~\cite{li2023revisiting}), we adopt RevIN, the instance normalization (inverse normalization) method \cite{kim2022reversible}, for input (output) samples to address the distribution shift in LTTP.

Similar to other recent LTTP methods (e.g., PatchTST~\cite{2023_nie_patchtst}, RLinear and RMLP~\cite{li2023revisiting}), we firstly adopt RevIN, an instance normalization (inverse normalization) method \cite{kim2022reversible}, for the data preprocess (postprocess) of input (output) to address the distribution shift in LTTP.
Secondly, we utilize the token embedding~\cite{2021_zhou_informer} to embed \tm~and \sm~into a latent space of \emph{d}-dimension and add positional information by the positional encoding~\cite{vaswani2017attention}.
We denote the obtained vectors of \tm~and \sm~after these processes as $E_{TM}$ and $E_{SM}$, respectively.
For temporal features, we propose an attention embedding layer to transform different types of temporal features to the \emph{d}-dimension of the latent space as well as exploiting the correlations among them.
We denote the embedded vectors of temporal features as $E_T$.
Thirdly, $E_{TM}, E_{SM}$ and $E_T$ are fed to \emph{C} layers of cross-modality fusion to: (1) learn the high-level representation of \tm~by a masked multi-head self-attention module, and (2) capture temporal correlations of \tm~and \sm~contributing to the prediction of \tm~by a masked multi-head temporal attention module. 
Finally, the output of the multi-layer cross-modality fusion is passed to a linear layer to project from the \emph{d}-dimension to the dimension of output samples. In addition, we utilize the idea of a multi-resolution recursive training strategy~\cite{niu2022mu2rest} to enhance the prediction performance. The details of the mentioned components are presented in this section.

Notably, our proposed architecture is similar to the decoder of a conventional Transformer model. 
Recently, encoder-only LTTP models such as the single linear models~\cite{li2023revisiting} and PatchTST \cite{2023_nie_patchtst} have shown superiority against the complex conventional encoder-decoder Transformer-based models. 
Inspired by them, our model uses the masked self-attention of the Transformer decoder to force the model to learn the backward correlations by only looking into the past data for each time step.

\subsection{Attention-based Temporal Embedding}
Temporal features such as month, date, hour, minute, day of week, and holiday are important factors for traffic prediction due to the periodicity in traffic patterns.
Previous works~\cite{2021_zhou_informer}~\cite{2021_wu_autoformer}~\cite{2022_zhou_fedformer} incorporated them to traffic prediction models by calculating the summation of linearly embedded vectors of them, then aggregating the summation to the target modality.

In our work, we propose an attention-based temporal embedding layer, employing a multi-head self-attention layer~\cite{vaswani2017attention} to capture the correlations among different types of temporal features as shown in Fig.~\ref{fig:overview}(b).
Given a set of temporal features $\{M_i, d_i, h_i, m_i, W_i, H_i\}$ at a timestamp \emph{i}-th, $M_i\in[1,12], d_i\in[1,31], h_i\in[0,23]$, $m_i\in[0,(60/R)-1]$ (used only if $R<60$ minutes), $W_i\in[0,6], H_i\in\{0,1\}$ are month, date, hour, minute, day of week, and holiday (0: normal day, 1: holiday), respectively. We embedded and combined them as follows:
\begin{equation}
    E_T = \sum Self\_Attention\_Layer(L_T)
\end{equation}
where $L_T = [E_{M_i}, E_{d_i}, E_{h_i}, E_{m_i}, E_{W_i}, E_{H_i}]$ and those element vectors are embedded vectors of $M_i, d_i, h_i, m_i, W_i$, and $H_i$ via separated linear embedding layers, respectively.

\subsection{Cross-Modality Fusion Layer}

\begin{figure*}[t]
    \centering
    \includegraphics[scale=0.33]{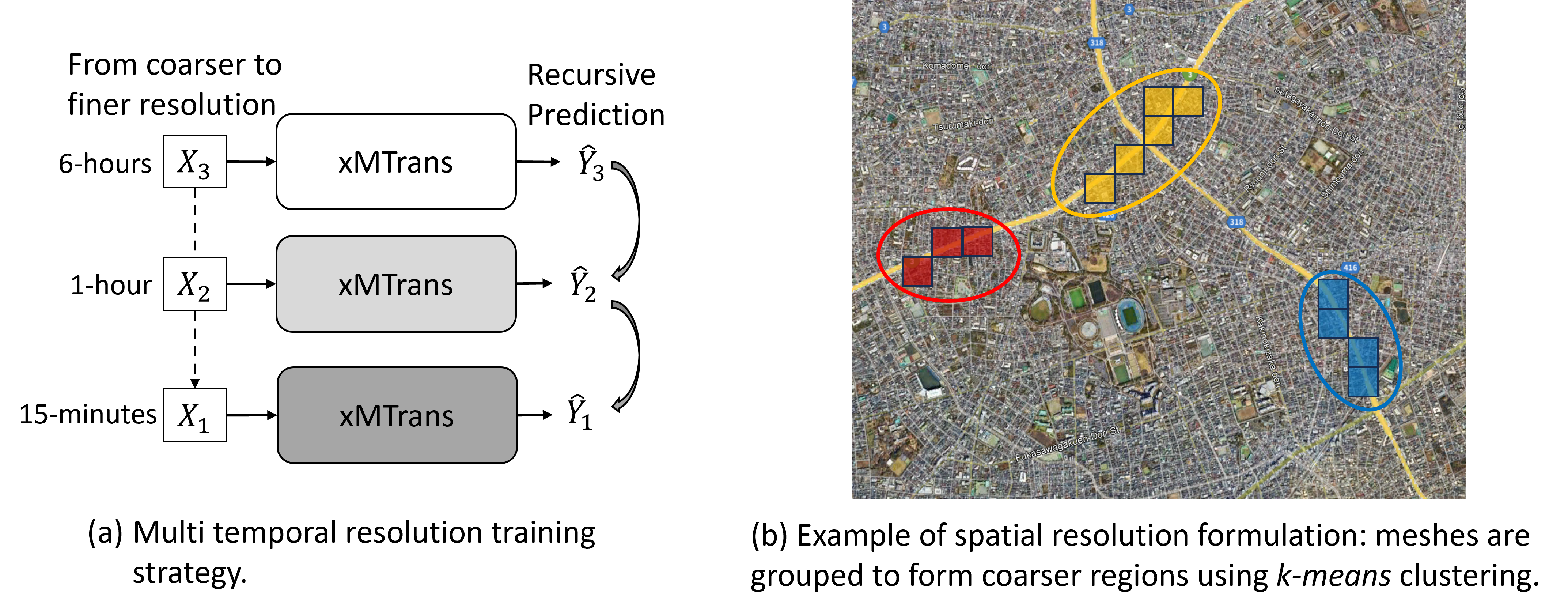}
    \caption{Illustration of temporal multi-resolution training strategy and spatial multi-resolution formulation.}
    \label{fig:recursive}
\end{figure*}

\subsubsection{Masked Multi-head Self-attention Module}
To learn the high-level representation of \tm, we use multi-head self-attention module~\cite{vaswani2017attention} as follows:
\begin{equation}
    Attention(Q,K,V) = Softmax(\frac{QK^T + Mask}{\sqrt{d_k}})V
\end{equation}
where $Q, K$ and $V$ are formed by $E_{TM}$; $d_k$ is the dimension of $K$; $Mask$ is a matrix to restrict the model to attend the right side of a current timestamp. 
We also add a residual connection to preserve the lower-level representation of \tm.

\subsubsection{Masked Multi-head Temporal Attention Module}
There could be temporal correlations in multi-modal traffic modelling. 
For instance, if we consider traffic congestion and rainfall amount, heavy rain occurs in a region, then traffic participants slow down their vehicles to avoid incidents due to limited visibility and slippery roads.
This reduction of speed is gradual, which might cause congestion after \emph{c} minutes, especially during rush hours.
The congestion at a timestamp \emph{t}-th and the rainfall amount at the timestamp \emph{(t-c)}-th reflects the temporal correlation between the two modalities.

Inspired by this kinds of situations, our model aims to learn the temporal correlations using a masked multi-head temporal attention. 
This layer is a variant of the masked multi-head self-attention layer, where the pairs of key and query are formed by the embedded temporal features $E_T$ while values are formed by the embedded vector $E_{SM}$ of \sm.
In this sense, the correlated values of time-based keys matched a time-based query are retrieved and combined to the representation of \tm~corresponding to that query.

After combining the representations of \tm~and \sm, the obtained combination is fed to the layer normalization~\cite{vaswani2017attention} and one-dimensional convolution like~\cite{2021_zhou_informer}.
The output of them is then assigned to $E_{TM}$ and input to the next cross-modality fusion layer to learn the higher-level representations and combinations. 

\subsection{Multi-Resolution Recursive Training Strategy}
Training a MTSP model with coarse and fine temporal-spatial resolutions presented the effectiveness in the Mu2ReST model~\cite{niu2022mu2rest}.
Here, we apply temporal multi-resolutions (TMR) and spatial multi-resolutions (SMR) for our \emph{xMTrans}.

\subsubsection{Temporal Multi-Resolutions}
Considering three temporal resolutions as illustrated in Fig.~\ref{fig:recursive}a,
$\{X_r, X'_r, \hat{Y}_r, Y_r\}$ is the set of \tm~input, \sm~input, prediction results, and ground truth regarding the resolution $r$ with $r \in \{r_1,r_2,r_3\}$ (e.g., \{15-minutes, 1-hour, 6-hours\}). 
The finest resolution of $r = r_1$ is our main target. 
We first train our \emph{xMTrans} with the inputs of $X_{r_3}$ and $X'_{r_3}$, then obtain the prediction $\hat{Y}_{r_3}$ using the loss function $\mathcal{L}(Y_{r_3}, \hat{Y}_{r_3})$, where $\mathcal{L}$ is the Mean Square Error (MSE) in this paper. 
Secondly, we train another \emph{xMTrans} with the inputs of $X_{r_{2}}$ and $X'_{r_{2}}$, then obtain $\hat{Y}_{r_{2}}$ using the loss function $\mathcal{L}(Y_{r_{2}}, \hat{Y}_{r_{2}}) + \mathcal{L}(\hat{Y}_{r_{2}\rightarrow r_{3}}, \hat{Y}_{r_{3}})$.
Here, our model ensures that the aggregation $\hat{Y}_{r_2\rightarrow r_3}$ of $\hat{Y}_{r_2}$ to the next higher resolution is close to the prediction made with the corresponding model $\hat{Y}_{r_3}$. 
Finally, we train \emph{xMTrans} with the inputs of $X_1$ and $X'_1$, obtains $\hat{Y}_1$ using the loss function $\mathcal{L}(Y_{r_1}, \hat{Y}_{r_1}) + \mathcal{L}(\hat{Y}_{r_{1}\rightarrow r_{2}}, \hat{Y}_{r_2})$, where $\hat{Y}_{r_{1}\rightarrow r_{2}}$ is the aggregation of $\hat{Y}_{r_1}$ to the next higher resolution.

\subsubsection{Spatial Multi-Resolutions}
In traffic modelling, neighbour regions often share the same characterises. 
Training MTSP models with fine and coarse spatial resolutions 
simultaneously can help the model to better capture those characterises while reducing the impact of data perturbations on the prediction model.
Here, we group nearby regions together to form ``coarse'' regions and aggregate time-series samples of the fine regions to form the samples of the coarse regions. 
Fig.~\ref{fig:recursive}b illustrates an example of grouping square meshes (covering a road with occurrences of traffic congestion) into three coarse regions, i.e., red, yellow, and blue regions. 
We apply the \textit{K-means} clustering to group those meshes based on geometric coordination of meshes' centroids.
In the training phase, we train \emph{xMTrans} by the samples of both fine and coarse regions to enhance the prediction performance. 
For the testing phase, we only evaluate our model's predictions on the fine regions.

\begin{table}[htbp]
    \centering
    \resizebox{0.48\textwidth}{!}{
    \begin{tabular}{ccc}  
         \hline
         Prediction task &TM& SM\\ 
         \hline
         Traffic congestion &Traffic congestion length (TCL)& People-flow (PF)\\ 
         Taxi demand&Green taxi& Yellow taxi\\ 
         \hline
    \end{tabular}
    }
    \caption{Description for settings of TM and SM in our experiments.}
    \label{tab:tm_sm}
\end{table}

\begin{table*}[htbp]
    \centering
    % \resizebox{9cm}{!}{
    \begin{tabular}{l|r|r|r|r|r|r}
        \hline
         Dataset&  \multicolumn{3}{c|}{TCL\&PF}& \multicolumn{3}{c}{NYT Green\&Yellow}\\
 & Train& Valid& Test& Train& Valid&Test\\ \hline
         Start Time&  2017-01-01 & 2018-01-01&2019-01-01& 2017-04-01 & 2017-05-01&2017-06-01\\
         End Time&  2017-06-30 & 2018-06-30&2019-06-30& 2017-04-30& 2017-05-31&2017-06-30\\ \hline
         Time interval&  \multicolumn{3}{c|}{15 minutes }& \multicolumn{3}{c}{30 minutes}\\ \hline
 \#Meshes& \multicolumn{3}{c|}{264}&\multicolumn{3}{c}{1544}\\ \hline
 \#Coarse regions&  \multicolumn{3}{c|}{64}&\multicolumn{3}{c}{365}\\ \hline
 Mesh size& \multicolumn{3}{c|}{250m $\times$ 250m}&\multicolumn{3}{c}{420m $\times$ 720m}\\ \hline
    \end{tabular}
    % }
    \caption{Description of training, validation, and testing sets for traffic congestion length and taxi demand predictions.}
    \label{tab:data}
\end{table*}

\section{Evaluations}
We evaluated our proposed method on the traffic congestion and taxi demand predictions. 
For traffic congestion prediction, we use a private dataset of traffic congestion length (denoted as TCL) as \tm~and a private dataset of people-flow (denoted as PF) as \sm.
These two datasets, however, will be published to research community via the DCCS platform according to~\cite{dao2023fostering}.
For taxi demand prediction, we use the New York Taxi (denoted as NYT) dataset, consisting of green and yellow taxis. Green taxi data is used as \tm~while yellow taxi plays a role as \sm. 
The overview setting for TM and SM is described in Table~\ref{tab:tm_sm}.

\subsection{Dataset}
\subsubsection{Traffic Congestion Length Dataset (TCL)}
JARTIC\footnote{https://www.jartic.or.jp/} provides this dataset through ``Congestion Statistics System''~\cite{jartic}.
Sensors that measure the speed of passing vehicles are deployed along general roads in Japan.
``Congestion'' is considered when the speed of passing vehicles is lower than $20km/h$. 
The length of the congestion is then computed according to the information captured by the sensors.
This dataset provides the minimum, average and maximum length of congestion (in meters) for every 5-minutes interval in each $250m\times250m$ cell-grid.

\subsubsection{People-Flow Dataset (PF)}
ZENRIN DataCom CO., LTD.\footnote{https://www.zenrin-datacom.net/en/} collects individual location data from mobile phone users who have given their consent through applications such as the ``Zenrin map navi'' service~\cite{docomonavi}. 
These data are collectively and statistically processed to conceal users' private information.
This dataset is named ``Konzatsu-Tokei (R) [Zenrin map navi]''~\cite{peopleflow}.
Original location data consists of GPS coordinates (latitude and longitude) transmitted at intervals of 1 hour and does not include any personally identifiable information.
This dataset provides the minimum, average, and maximum number of people appearing in each $250m\times250m$ cell-grid, for every 1-hour interval.

\subsubsection{New York Taxi Dataset (NYT)}
New York City Taxi \& Limousine Commission~\footnote{https://www1.nyc.gov/site/tlc/about/tlc-trip-record-data.page} provides the trip records of yellow and green taxis including pick-up and drop-off dates/times, locations, and other kinds of information. Following the work of~\cite{ye2019co}, we use the data from April 1st, 2016 to June 30th, 2016, and divide the New York City by 720m × 420m rectangles. We aggregated the yellow and green taxi demands (the numbers of pick-up records) respectively for every half hour per rectangle. We denote this dataset as NYT.

\subsection{Experimental Settings}
For our proposed \emph{xMTrans}, we set the number of cross-modality fusion layers $C=2$. 
The number of heads in the attention layers was 8 and the dimension of the latent space $d$ was 512. 
We used Adam optimization method to train \emph{xMTrans}. MSE was used to compute the loss between the groundtruth and the predicted values.

For both traffic congestion and green taxi demand predictions, our model look-backed 24-hours to predict 12-hours ahead. 
Here, we evaluated our model by Mean Absolute Error (MAE) and Root Mean Square Error (RMSE) in the range of 3-hours (short term), 6-hours (intermediate term), and 12-hours (long term) ahead. 
The set of resolutions in TMR $\{r_1, r_2, r_3\}$ were set as \{15-minutes, 1-hour, 6-hours\} and \{30-minutes, 2-hours, 6-hours\} for traffic congestion and green taxi demand predictions, respectively.

In addition, for the traffic congestion prediction, we used records of TCL and PF from the Setagaya Ward of Tokyo in our experiments. 
We used the average values of traffic congestion length and people-flow to predict the average the congestion length.
Since TCL and PF datasets have different time intervals, we made both of them consistent with a 15-minute interval. 
Specifically, we aggregated the TCL data from 5-minute into 15-minute time interval and converted the PF data from 1-hour to 15-minute interval by assigning the value at every 15-minute within 1-hour as the value of 1-hour. 
The detailed training, validation, and testing sets for traffic congestion and green taxi demand predictions are provided in Table~\ref{tab:data}.

\begin{table*}[t]
    \centering
    \resizebox{16cm}{!}{
    \begin{tabular}{l|rr|rr|rr|rr|rr|rr} 
         \hline
         &  \multicolumn{6}{c|}{TCL}& \multicolumn{6}{c}{NYT Green}\\
         &  \multicolumn{2}{c}{3-hours}&  \multicolumn{2}{c}{6-hours}&  \multicolumn{2}{c|}{12-hours}& \multicolumn{2}{c}{3-hours}& \multicolumn{2}{c}{6-hours}& \multicolumn{2}{c}{12-hours}\\ \hline
         Methods&  RMSE&  MAE&  RMSE&  MAE&  RMSE& MAE& RMSE& MAE& RMSE& MAE& RMSE&MAE\\ 
         % HA~\cite{li2017diffusion} & 181.08 & 86.80 & 181.15 & 86.85 & 181.26 & 86.95 & 1.81 & 0.49 & 1.81 & 0.49 & 1.81 & 0.49\\ 
         %        & (-) & (-) & (-) & (-) & (-) & (-) & (-) & (-) & (-) & (-) & (-) & (-) \\
         RLinear~\cite{li2023revisiting}& \underline{141.14} & \underline{77.23} & \underline{147.82} & \underline{80.52} & 152.45 & 83.22 & 1.19& \underline{0.38} & 1.23 & \underline{0.39} & 1.27& \underline{0.39}\\ 
                & \underline{(0.18)} & \underline{(0.40)} & \underline{(0.11)} & \underline{(0.31)} & (0.09) & (0.30) & (0.01)& \underline{(0.00)} & (0.02)& \underline{(0.00)}& (0.02)& \underline{(0.00)}\\
        RMLP~\cite{li2023revisiting}& 144.66 & 78.92 & 150.49 & 81.53 & 153.97 & 83.82 & \textbf{1.12} & \textbf{0.37} & \underline{1.23} & \underline{0.39} & 1.27 & \underline{0.39} \\ 
        & (0.43) & (0.14) & (0.40) & (0.31) & (0.24) & (0.08) & \textbf{(0.00)} & \textbf{(0.00)} & \underline{(0.00)} & \underline{(0.00)} & (0.00) & \underline{(0.00)}\\
         Autoformer~\cite{2021_wu_autoformer}& 154.50 & 90.09 & 154.44 & 89.84 & 156.77 & 91.88 & 1.56 & 0.46 & 1.55 & 0.47& 1.57 &0.48\\
                    & (1.08) & (1.99) & (2.24) & (3.04) & (2.25) & (2.94) & (0.06)& (0.03)& (0.06)& (0.03)& (0.05)&(0.02)\\
         Autoformer*& 156.26 & 91.89 & 156.25 & 91.29 & 157.28 & 91.64 & 1.58 & 0.47 & 1.53 & 0.46 & 1.57 &0.48 \\
                    & (070) & (0.64) & (1.08) & (0.13) & (1.17) & (0.30) & (0.06)& (0.01)& (0.02)& (0.01)& (0.02)&(0.01)\\
         Pyraformer~\cite{2022_liu_pyraformer}& 172.20 & 101.48 & 179.28 & 105.55 & 182.96 & 108.97 & 1.45 & 0.54 & 1.48 & 0.53 & 1.53 &0.54 \\ 
                   & (6.02) & (4.94) & (2.63) & (2.99) & (2.52) & (1.75) & (0.02)& (0.00)& (0.01)& (0.00)& (0.01)&(0.00)\\
         Pyraformer*& 172.57 & 101.95 & 179.43 & 105.90 & 181.77 & 107.15 & 1.37 & 0.51 & 1.41 & 0.51 & 1.46& 0.52 \\ 
                    & (4.43) & (4.73) & (4.86) & (5.49) & (4.00) & (4.42) & (0.02)& (0.00)& (0.02)& (0.01)& (0.01)& (0.00)\\
         FEDformer~\cite{2022_zhou_fedformer}& 162.90 & 96.73 & 160.91 & 94.14 & 160.79 & 92.65 & 1.27 & 0.49 & 1.26 & 0.48 & 1.30 &0.49 \\
                    & (1.61) &  (0.92) & (0.94) & (0.39) & (1.27) &  (0.23) & (0.03)& (0.01)& (0.03)& (0.01)& (0.03)&(0.01)\\ 
        FEDformer*& 174.18 & 104.85 & 172.55 & 102.30& 172.16 & 100.58 & 1.24 & 0.48 & 1.23 & 0.47 & 1.27 &0.48 \\
                    & (2.05)&  (1.99)& (1.64)& (1.45)& (1.58)&  (1.35)& (0.03)& (0.00)& (0.01)& (0.00)& (0.02)&(0.00)\\ 
        PatchTST~\cite{2023_nie_patchtst}& 142.38 & 79.02 & 148.33 & 81.02 & \underline{151.87} & \underline{83.17} & 1.14 & \underline{0.38} & 1.23 & 0.39 & \underline{1.26} &0.40 \\
                 & (0.64) & (0.67) & (0.42) & (0.44) & \underline{(0.17)} & \underline{(0.22)} & (0.00) & \underline{(0.00)}& (0.02)& (0.00)& \underline{(0.02)} &(0.00)\\
        % iTransformer~\cite{liu2023itransformer}& 142.54 & \underline{74.62} & \underline{150.35} & \underline{78.59} & 153.49 & \underline{81.84} & \underline{1.13} & \textbf{0.36} & \underline{1.23} & \underline{0.38} & 1.28 & \underline{0.39}\\
                 % & (0.48) & \underline{(0.24)} & \underline{(0.62)} & \underline{(0.26)} & (0.28) & \underline{(0.44)} & \underline{(0.01)} & \textbf{(0.00)} & \underline{(0.00)} & \underline{(0.00)}& (0.00)& \underline{(0.00)}\\
        % MegaCRN~\cite{jiang2023spatio} &  165.26 & 80.70 & 188.59 & 91.43 & 205.46 & 100.09 &
        %                 1.50& 0.39 & 1.58 & 0.40 & 1.66 & 0.41\\ 
        %            & (3.20)& (0.87)& (2.94) & (0.62) & (2.69) & (0.09) &
        %                 (0.06)& (0.04)& (0.09) & (0.03) & (0.15) & (0.04)\\
        3D-CNN~\cite{3DCNN} & 157.63 & 111.14 & 157.08 & 110.74 & 168.28 & 121.91 & -
                        & - & - & - & - & -\\ 
                   & 0.48 & 0.28 & 0.15 & 0.15 & 1.11 & 0.79
                        & (-) & (-) & (-) & (-) & (-) & (-)\\
         Ours: \emph{xMTrans}& \textbf{135.33}& \textbf{71.57}& \textbf{136.73}& \textbf{72.51}& \textbf{138.57}& \textbf{73.69} & \underline{1.14} & \textbf{0.37} & \textbf{1.14} & \textbf{0.37} & \textbf{1.16} & \textbf{0.37} \\
                 &  \textbf{ (0.67)}& \textbf{(0.28)} & \textbf{(0.74)} & \textbf{(0.41)} & \textbf{(0.90)} & \textbf{(0.45)} & \underline{(0.01)} & \textbf{(0.00)} & \textbf{(0.01)} & \textbf{(0.00)} & \textbf{(0.01)}& \textbf{(0.00)}\\ \hline
    \end{tabular}
    }
    \caption{Comparison to baseline methods. Results in Bold (Underline) are the best (second-best) values. Methods with ``*" indicate multi-modal models using feature concatenation. Results are in the format: Average (standard deviation).}
    \label{tab:compare_to_SOTA}
\end{table*}

\begin{table*}[t]
    \centering
    \resizebox{18cm}{!}{
    \begin{tabular}{l|l|rr|rr|rr|rr|rr|rr} 
         \hline
          &&  \multicolumn{6}{c|}{TCL}& \multicolumn{6}{c}{NYT Green}\\
          &&  \multicolumn{2}{c}{3-hours}&  \multicolumn{2}{c}{6-hours}&  \multicolumn{2}{c|}{12-hours}& \multicolumn{2}{c}{3-hours}& \multicolumn{2}{c}{6-hours}& \multicolumn{2}{c}{12-hours}\\ \hline
          Exp.&$[Query\_Key\_Value]$&  RMSE&  MAE&  RMSE&  MAE&  RMSE& MAE& RMSE& MAE& RMSE& MAE& RMSE&MAE\\ \hline
          E1&$[TM\_TM\_TM]_1$& 155.58 & 84.58 & 154.04 & 83.01 & 156.54 & 84.21 
& 1.20 & \underline{0.38} & 1.21 & \underline{0.38} & 1.23 & \underline{0.38} 
\\
                     && (2.11)& (1.69)& (0.49)& (1.36)& (0.03)& (0.72)
& (0.01)& \underline{(0.00)}& (0.01)& \underline{(0.00)}& (0.00)&\underline{(0.00)}\\
          E2&\hspace{4 mm} $+ [TM\_TM\_TM]_2$ & 156.47 & 84.31 & 154.80 & 82.52 & 158.37 & 84.79 & \underline{1.19}& \underline{0.38}& 1.20& \underline{0.38}& 1.22& \underline{0.38}\\
                     && (3.08)& (2.04)& (0.67)& (0.91)& (0.42)& (0.38)& \underline{(0.01)}& 
\underline{(0.00)}& (0.01)& \underline{(0.00)}& (0.00)& \underline{(0.00)}\\
          E3&\hspace{4 mm} $+ [TM\_TM\_SM]_2$ & \underline{150.77} & \underline{79.73} & \underline{152.25} & \underline{80.11} & \underline{156.57} & \underline{82.68} & \underline{1.19} & \underline{0.38} & \underline{1.19} & \underline{0.38} & \underline{1.21} & \underline{0.38} \\
                     && \underline{(1.58)}& \underline{(1.05)}& \underline{(0.59)}& \underline{(0.63)}& \underline{(0.26)}& \underline{(0.35)}& \underline{(0.01)}& \underline{(0.00)}& \underline{(0.00)}& \underline{(0.00)}& \underline{(0.00)}&\underline{(0.00)}\\
         % \hspace{4 mm} $+ [TM\_T\_SM]_2$ & 144.43 & 75.72 & 146.59 & 76.71 & 150.19 & 79.05 & 1.18 & 0.38 & 1.18 & 0.38 & 1.20 &0.38 \\
         %            & (0.99)& (0.20)& (1.58)& (0.55)& (1.00)& (0.15)& (0.00)& (0.00)& (0.00)& (0.00)& (0.00)&(0.00)\\
          E4&\hspace{4 mm} $ + [T\_T\_SM]_2\ (\emph{xMTrans})$& \textbf{135.33}& \textbf{71.57}& \textbf{136.73}& \textbf{72.51}& \textbf{138.57}& \textbf{73.69} & \textbf{1.14}  & \textbf{0.37} & \textbf{1.14} & \textbf{0.37} & \textbf{1.16} & \textbf{0.37} \\
            &&  \textbf{ (0.67)}& \textbf{(0.28)} & \textbf{(0.74)} & \textbf{(0.41)} & \textbf{(0.90)} & \textbf{(0.45)} & \textbf{(0.01)} & \textbf{(0.00)} & \textbf{(0.01)} & \textbf{(0.00)} & \textbf{(0.01)}& \textbf{(0.00)}\\ \hline
    \end{tabular}
    }
    \caption{Ablation study for query, key, and values in the cross-modality fusion layer. \textit{TM}, \textit{SM}, and \textit{T} stand for target modality, support modality and temporal features. $[~]_1$ and $[~]_2$ stand for the masked multi-head self-attention and the masked multi-head temporal attention modules, respectively.}
    \label{tab:ablation_study_qkv}
\end{table*}

\subsection{Baselines}
We compared our \emph{xMTrans} to LTTP models: RLinear~\cite{li2023revisiting}, RMLP~\cite{li2023revisiting},
Autoformer~\cite{2021_wu_autoformer}, Pyraformer~\cite{2022_liu_pyraformer}, FEDformer~\cite{2022_zhou_fedformer}, and PatchTST~\cite{2023_nie_patchtst}.
% These methods are mentioned in the section of related work. 
We basically re-used the same settings of these original implementation and fine-tuned some parameters to achieve well performance on our datasets.  
For Pyraformer, we used a sliding window size of 2 when building up the pyramid. 
For PatchTST, we applied the patch length of 16 and set the stride value of 8 when patching the time-series input.
We conducted experiments using GPUs P100 with 16GB memory for all models excepting PatchTST, for which we partially employed an A100 GPU with 80GB memory.

Since those mentioned models are originally designed for uni-modal LTTP, we slightly modified Pyraformer, Autoformer and FEDformer to enable them to be applicable for multi-modal LTTP by simply concatenating \tm~and \sm~in the feature dimension. 
Those modified methods are marked as ``Method*''.
Note that only those methods allowing freedom of the feature size can be applied for our minor modification.

For traffic congestion prediction, we additionally implemented 3D-CNN method of~\cite{3DCNN} for comparison using the same settings.

\subsection{Main Results}
To examine the robustness of the prediction models, we repeated each experiment three times with different initialization seeds and calculated the average and standard deviation values of RMSE and MAE obtained. 

Table~\ref{tab:compare_to_SOTA} presents the comparisons of our proposed model and mentioned baseline models. 
The results indicate the superiority of our proposed model \emph{xMTrans} to previous methods on long-term traffic congestion and taxi demand predictions.
RLinear, RMLP, and PatchTST, which are recent three state-of-the-art methods, achieve the second best performance in several cases.
Regarding the short-term taxi demand prediction, RMLP performs slightly better than our \emph{xMTrans} in terms of RMSE.

It is worth to note that Method* performs worse than Method on TCL even they consumed SM for their predictions. 
It shows that simply concatenating TM and SM could degrade the performance because of mixing two modalities without properly modelling correlations between them.

\begin{table*}[t]
    \centering
    \resizebox{16cm}{!}{
    \begin{tabular}{l|rr|rr|rr|rr|rr|rr} 
         \hline
         &  \multicolumn{6}{c|}{TCL}& \multicolumn{6}{c}{NYT Green}\\
         &  \multicolumn{2}{c}{3-hours}&  \multicolumn{2}{c}{6-hours}&  \multicolumn{2}{c|}{12-hours}& \multicolumn{2}{c}{3-hours}& \multicolumn{2}{c}{6-hours}& \multicolumn{2}{c}{12-hours}\\ \hline
         Method&  RMSE&  MAE&  RMSE&  MAE&  RMSE& MAE& RMSE& MAE& RMSE& MAE& RMSE&MAE\\ 
         W/o SMR& \underline{136.82} & \underline{72.04} & \underline{138.10} & \underline{72.96} & \underline{139.90} & \underline{74.23} & 1.16 & 0.37 & 1.16 & 0.37 & 1.18 &0.37 \\
                    & \underline{(0.58)}& \underline{(0.52)}& \underline{(0.78)}& \underline{(0.68)}& \underline{(0.93)}& \underline{(0.63)}& (0.01)& (0.00)& (0.01)& (0.00)& (0.01)&(0.00)\\
         W/o TMR & 137.82 & 73.16 & 139.01 & 73.67 & 140.73 & 74.80 & \textbf{1.12} & \textbf{0.36} & \textbf{1.13} & \textbf{0.36} & \textbf{1.15} & \textbf{0.37} \\
                    & (1.23)& (0.39)& (1.09)& (0.29)& (1.01)& (0.27)& \textbf{(0.01)} & \textbf{(0.00)}& \textbf{(0.00)}& \textbf{(0.00)}& \textbf{(0.00)}& \textbf{(0.00)}\\
         W/o Self-attention in TE & 147.95 & 78.07 & 151.48 & 79.25 & 154.00 & 81.42 & 1.19 & 0.38 & 1.20 & 0.38 & 1.22 &0.38 \\
                    & (0.41)& (0.68)& (0.49)& (0.77)& (0.05)& (0.34)& (0.02)& (0.00)& (0.02)& (0.00)& (0.02)&(0.00)\\
         Ours: \emph{xMTrans} & \textbf{135.33}& \textbf{71.57}& \textbf{136.73}& \textbf{72.51}& \textbf{138.57}& \textbf{73.69} & \underline{1.14}  & \underline{0.37} & \underline{1.14} & \underline{0.37} & \underline{1.16} & \textbf{0.37} \\
          \vspace{0.4mm}  &  \textbf{ (0.67)}& \textbf{(0.28)} & \textbf{(0.74)} & \textbf{(0.41)} & \textbf{(0.90)} & \textbf{(0.45)} & \underline{(0.01)} & \underline{(0.00)} & \underline{(0.01)} & \underline{(0.00)} & \underline{(0.01)}& \textbf{(0.00)}\\ \hline
    \end{tabular}
    }
    \caption{Ablation study for other components.}
    \label{tab:ablation_study_others}
\end{table*}

\begin{figure*}[t]
	\centering
	\begin{subfigure}{0.43\linewidth}
		\includegraphics[width=\linewidth]{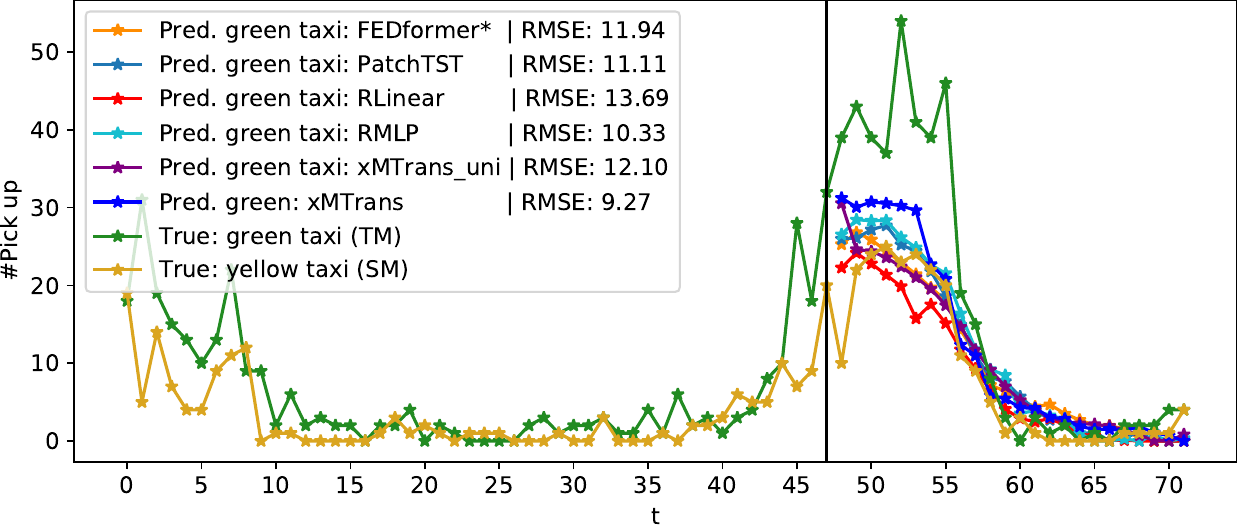}
		\caption{Friday $\rightarrow$ Saturday}
		\label{fig:examples_a}
	\end{subfigure} \hfill
	\begin{subfigure}{0.43\linewidth}
		\includegraphics[width=\linewidth]{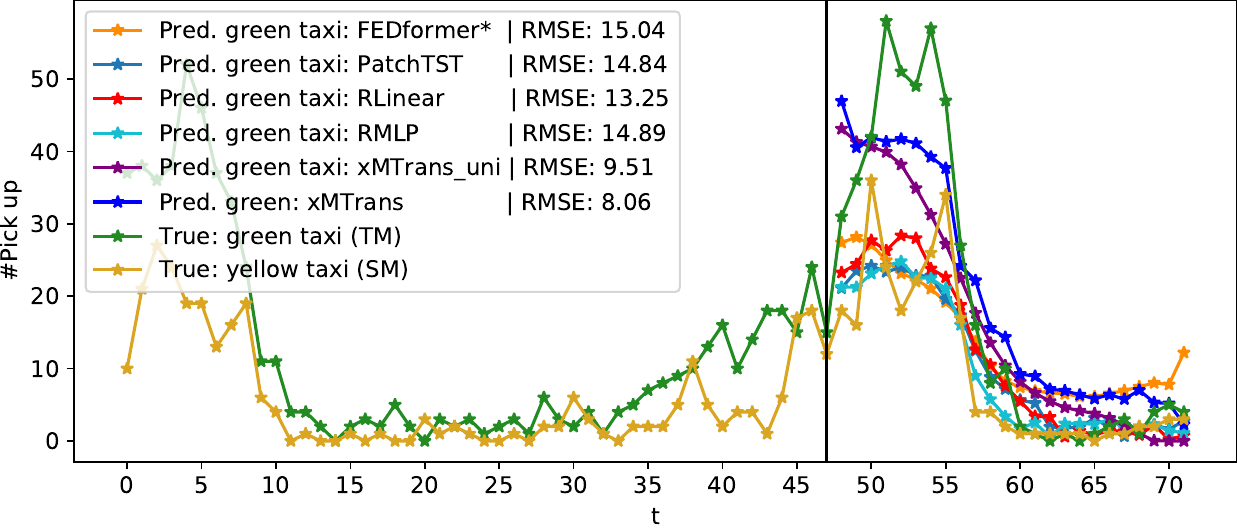}
		\caption{Saturday $\rightarrow$ Sunday}
		\label{fig:examples_b}
	\end{subfigure}\\
     \begin{subfigure}{0.43\linewidth}
    		\includegraphics[width=\linewidth]{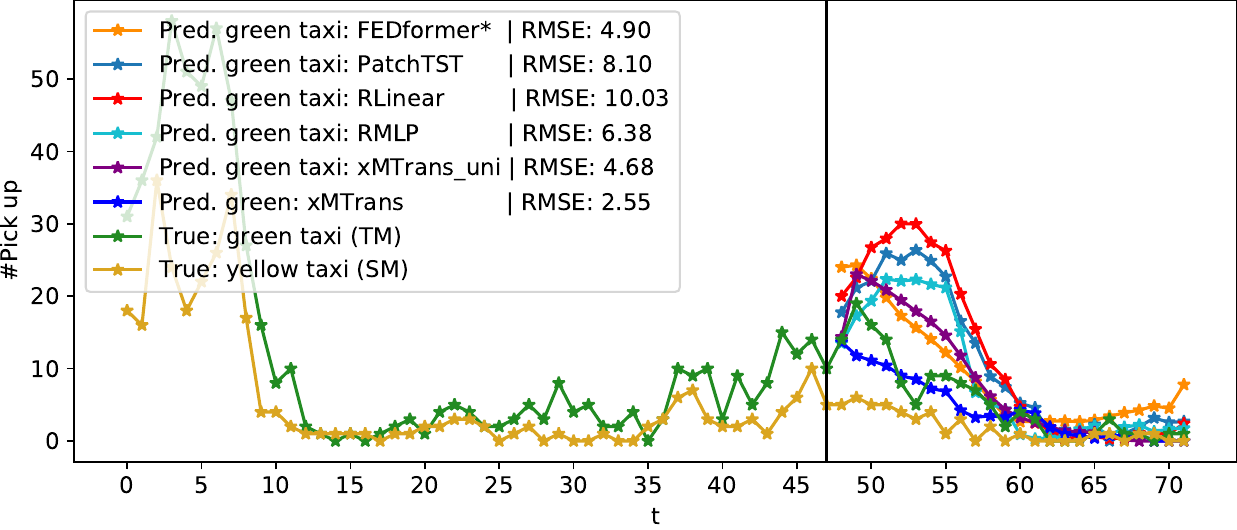}
    		\caption{Sunday $\rightarrow$ Monday}
    		\label{fig:examples_c}
    	\end{subfigure} \hfill
    	\begin{subfigure}{0.43\linewidth}
    		\includegraphics[width=\linewidth]{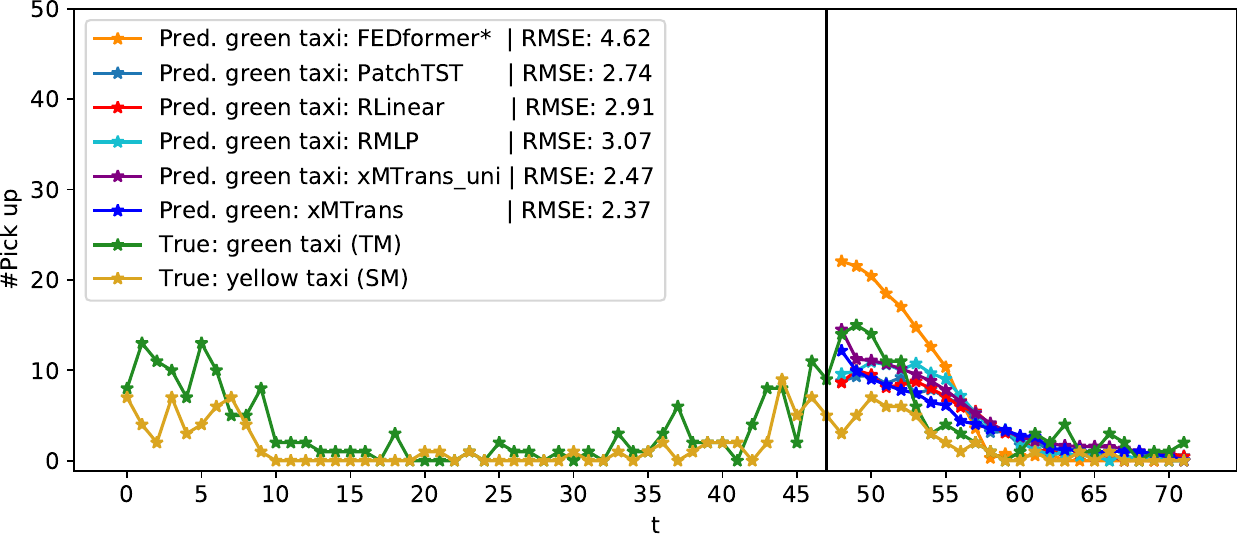}
    		\caption{Tuesday $\rightarrow$ Wednesday}
    		\label{fig:examples_d}
    	\end{subfigure}
	\caption{Examples of predictions from our \emph{xMTrans}, its uni-modal version, and the highlighted baselines. The predictions start from $(t=48)$. $(t=0)$ is a time step at 0:00, while $(t=48,71)$ are 0:00 and 11:45 of the next following day, respectively.}
	\label{fig:examples}
\end{figure*}

\begin{figure}[t]
	\centering
	\begin{subfigure}{0.4\linewidth}
		\includegraphics[width=\linewidth]{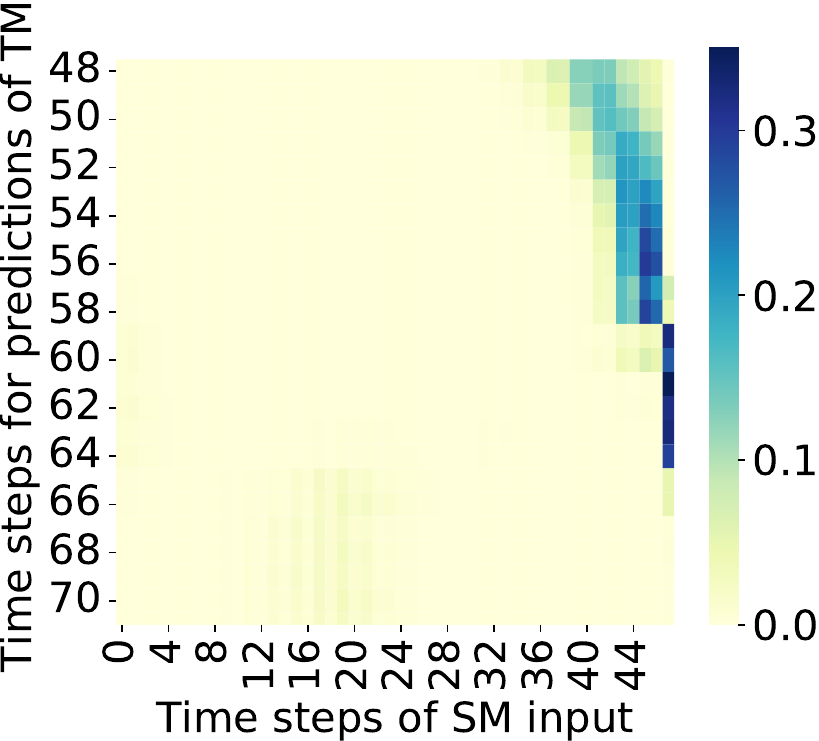}
		\caption{Friday $\rightarrow$ Saturday}
		\label{fig:attn_a}
	\end{subfigure}
    ~
	\begin{subfigure}{0.4\linewidth}
		\includegraphics[width=\linewidth]{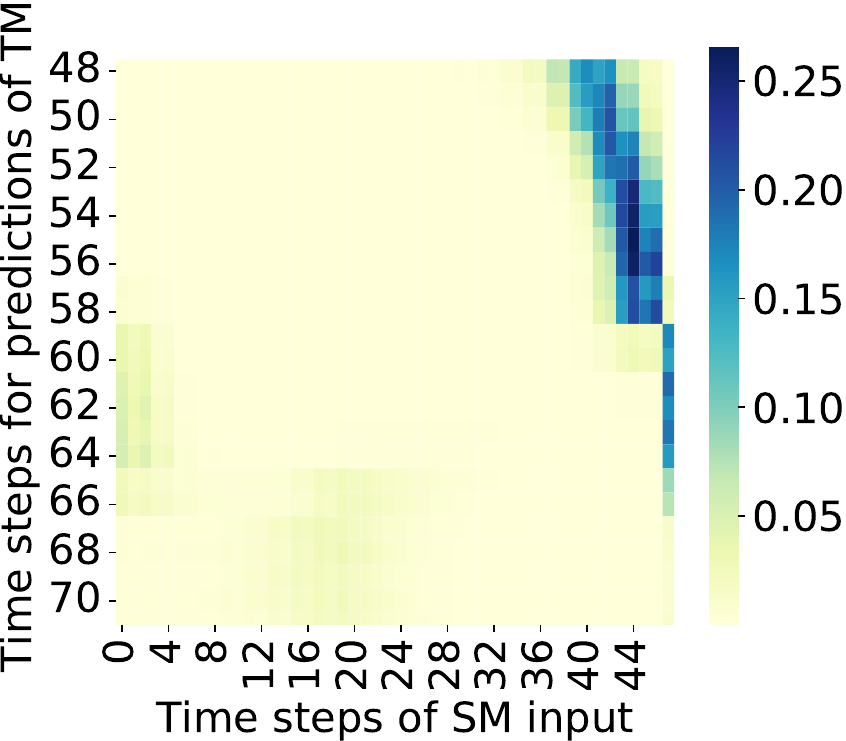}
		\caption{Saturday $\rightarrow$ Sunday}
		\label{fig:attn_b}
	\end{subfigure} \\
    \begin{subfigure}{0.4\linewidth}
		\includegraphics[width=\linewidth]{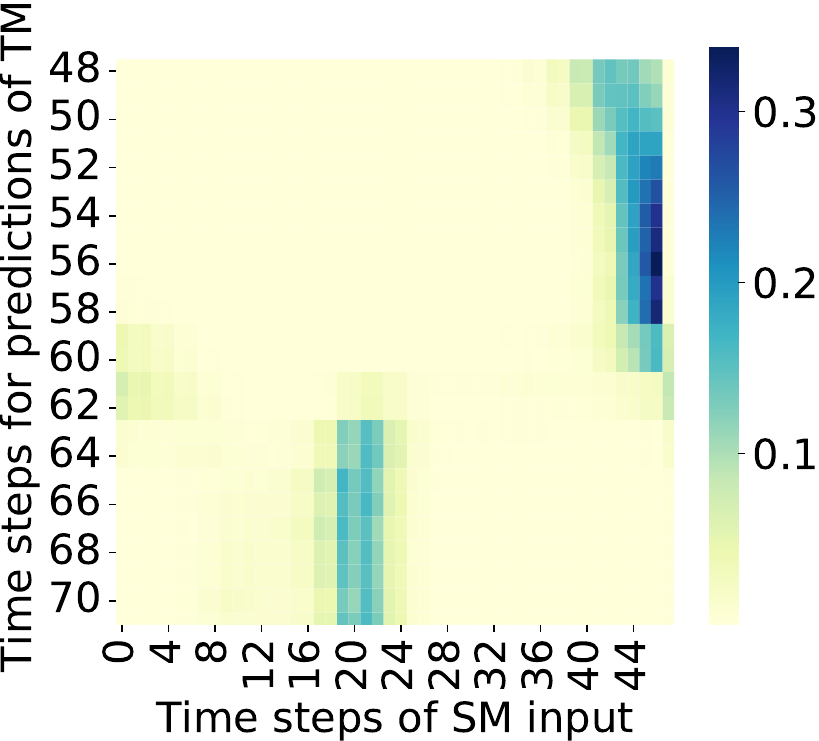}
		\caption{Sunday $\rightarrow$ Monday}
		\label{fig:attn_c}
	\end{subfigure}
    ~
	\begin{subfigure}{0.4\linewidth}
		\includegraphics[width=\linewidth]{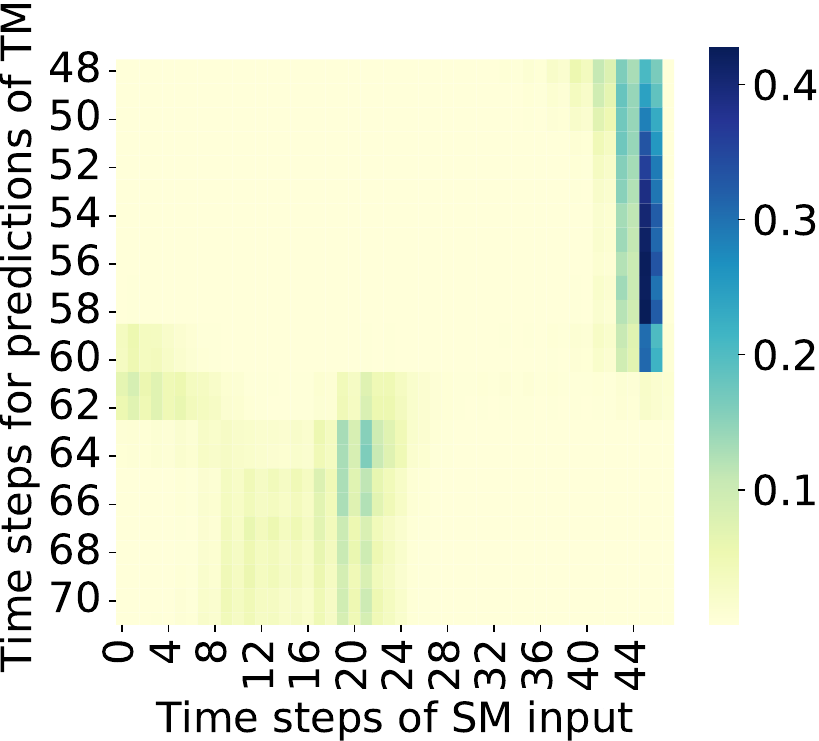}
		\caption{Tuesday $\rightarrow$ Wednesday}
		\label{fig:attn_d}
	\end{subfigure}
	\caption{Average maps of eight attention maps from the masked multi-head attention module in the last cross-modality fusion layer.}
	\label{fig:att_maps}
\end{figure}

% \begin{figure*}
%     \centering
%     \includegraphics[width=1.0\linewidth]{att_maps.pdf}
%     \caption{Eight attention maps of the masked multi-head attention module in the last cross-modality fusion layer.}
%     \label{fig:att_maps}
% \end{figure*}

\subsection{Ablation Study}
We comprehensively conducted ablation study for our proposed method.
First, we examine the effective of time-based keys and queries by forming the key and query base on \tm. 
Secondly, we remove \sm~and use \tm~to check the contribution of \sm~in our model. 
Table~\ref{tab:ablation_study_qkv} shows the results of these experiments.
Overall, the architecture of \emph{xMTrans} (E4) achieved the best performance among other variants.
Without using the masked multi-head temporal attention module (E1) results the worst performance.
Using \tm-based keys and queries (E3) significantly degrades the prediction performance.
Removing \sm~(E2) also causes the degradation, indicating the efficacy of \sm.

Furthermore, we conducted experiments to show the impacts of other components in our proposed model.
We sequentially tested SMR, TMR and the importance of self-attention layer in the temporal embedding (TE).
Table~\ref{tab:ablation_study_others} presents the results of these experiments.
Without SMR, our model's performance slightly decreases.
Without TMR, it also causes a degradation of the performance on TCL dataset.
In contrast, the performance of green taxi prediction becomes slightly better.
We will investigate this phenomenon in our future work. Without using the self-attention layer in TE significantly degrades the performance on both datasets.

\subsection{Prediction Analysis and Visualization}
This section analyzes and visualizes examples of results from typical prediction models. Since TCL and PL are private datasets, we only present examples from the NYT dataset.
In Fig.~\ref{fig:examples}, we plot four prediction samples of green taxi demand from 0:00 to 11:45, obtained from our \emph{xMTrans}, the typical baseline models (i.e., FEDformer*, RLinear, RMLP, and PatchTST based on Table~\ref{tab:compare_to_SOTA}), and the uni-modal version of our model (denoted as xMTrans\_uni) which is without the multi-head temporal attention module.
These samples are in the same location but on different dates.
Since the taxi demands on weekend and weekday are significantly different (especially at early morning), we show two prediction samples on Saturday (Fig.~\ref{fig:examples_a}) and Sunday (Fig.~\ref{fig:examples_b}); and two others on Monday (Fig.~\ref{fig:examples_c}) and Wednesday (Fig.~\ref{fig:examples_d}).
The values before $(t=48)$ are the input data to our model. 
Note that RLinear, RMLP, and PatchTST are uni-modal models so they only received the input of green taxi demand values.
We observe that our proposed model performs better than the others on all samples.
Especially, \emph{xMTrans} significantly well performs than others on Monday and Sunday at time steps from 48 to 60. 
In these cases, it seems that the other models are probably confused whether the prediction period is on Monday or Sunday because the input in those cases are quite similar as shown in Fig.~\ref{fig:examples_b} and Fig.~\ref{fig:examples_c}.
In fact, one correlation of two taxis to discriminate those cases is to look at the input's time steps from 35 to 47. 
We realized that the the demands of both taxis at that period increase slightly but lower than 15 before the weekday, while those before the weekend significantly increase to more than 20.

Our multi-modal model is able to capture this correlation to achieve better performance on both weekend and weekday predictions. 
To show this, we visualize average maps of eight attention maps from the masked multi-head temporal attention module in our last cross-modality fusion layer used for the predictions of the samples in Fig.~\ref{fig:examples}.
X-axis presents time steps of yellow taxi demands inputted while Y-axis describes time steps for predictions of green taxi demands.
These plots reveal that the predictions at time steps from 48 to around 60 pay attention on values of yellow taxi demands at time steps ranging from 35 to 47 as our expectation.
Besides, the predictions at time steps from 60 to 71 tend to look at time steps from 15 to 25, in which their trend and values are similar to the groundtruth of the prediction.
These points confirm the effectiveness of our \emph{xMTrans} in capturing temporal correlations among two modalities.

\section{Conclusion and Future Works}
This paper introduced \emph{xMTrans}, a novel temporal attentive cross-modality transformer model designed to explore 
%the correlations among the temporal dimension of two modalities, 
the temporal correlation between two modalities, 
i.e., a target modality (for prediction, denoted as TM), and a support modality (denoted as SM). 
We conducted experiments for our proposed model on long-term traffic congestion and
taxi demand predictions.
For the traffic congestion prediction, we used two private datasets of congestion length (as \tm) and people-flow data (as \sm).
For the taxi demand prediction, we utilized the New York taxi dataset, including green (as \tm) and yellow (as \sm) taxi demand records.
Our proposed method outperformed recent methods on these two predictions. 

For future work, we would like to investigate the methods for exploring both temporal and spatial correlations for long-term traffic predictions. 
Besides, we will extend our prediction model to utilize more than two modalities.
% Besides, we will design a architecture with a capability of leveraging advantages of large language models, which have shown impressive performance recently in the natural language processing field. 

\section*{Acknowledgment}
This research and development were partially carried out as the NICT advanced communication/broadcasting research and development commissioned research, No. 22701 (Research and development of next-generation IoT data utilization technology for highly sustainable action support).

\bibliographystyle{IEEEtran}
\bibliography{reference_mtsp}

\end{document}